\documentclass{article}

\usepackage{arxiv}

\usepackage[utf8]{inputenc} 
\usepackage[T1]{fontenc}    
\usepackage{hyperref}       
\usepackage{url}            
\usepackage{booktabs}       
\usepackage{amsfonts}       
\usepackage{nicefrac}       
\usepackage{microtype}      
\usepackage{lipsum}
\usepackage{graphicx}
\usepackage{color}
\usepackage{caption}
\usepackage{amsmath}
\usepackage{multirow} 
\usepackage{array, makecell}

\title{Independent Component Analysis for Trustworthy Cyberspace during
	High Impact Events: An Application to Covid-19}

\author{
	~~~~~~~~~~~Zois Boukouvalas\\
	~~~~~~~~~~~American University\\
	~~~~~~~~~~~Washington, DC 20016, USA 
	\And
	~~~~~~~~Christine Mallinson\\
	~~~~~~~~University of Maryland Baltimore County\\
	~~~~~~~~Baltimore, MD 21250, USA
	\AND
	~~~~~~~~Evan Crothers\\
	~~~~~~~~University of Ottawa\\
	~~~~~~~~Ottawa, ON, Canada
	\And
	~~~~~~~~~~~~~~~~Nathalie Japkowicz\\
	~~~~~~~~~~~~~~~~American University\\
	~~~~~~~~~~~~~~~~Washington, DC 20016, USA 
	\AND
	Aritran Piplai\\
	University of Maryland Baltimore County\\
	Baltimore, MD 21250, USA
	\And
	Sudip Mittal\\
	University of North Carolina Wilmington\\
	Wilmington, NC 28403, USA
	\AND
	Anupam Joshi\\
	University of Maryland Baltimore County\\
	Baltimore, MD 21250, USA
	\And
	T\"{u}lay Adal{\i} \\
	University of Maryland Baltimore County\\
	Baltimore, MD 21250, USA
}

\begin{document}
\maketitle

\begin{abstract}
Social media has become an important communication channel during high impact events, such as the COVID-19 pandemic. As misinformation in social media can rapidly spread, creating social unrest, curtailing the spread of misinformation during such events is a significant data challenge. While recent solutions that are based on machine learning have shown promise for the detection of misinformation, most widely used methods include approaches that rely on either handcrafted features that cannot be optimal for all scenarios, or those that are based on deep learning where the interpretation of the prediction results is not directly accessible. In this work, we propose a data-driven solution that is based on the ICA model, such that knowledge discovery and detection of misinformation are achieved jointly. To demonstrate the effectiveness of our method and compare its performance with deep learning methods, we developed a labeled COVID-19 Twitter dataset based on socio-linguistic criteria.
\end{abstract}


\section{Introduction}
With the evolution of social media technologies, there has been a fundamental change in how the information is accessed, shared, and propagated. Propagation of information, and in particular that of misinformation, becomes especially important during high impact events like natural disasters, terrorist attacks, periods of political transition or unrest, and financial instability. A recent example of a high impact event is the novel Coronavirus Disease 2019 (COVID-19) where misinformation is dangerously spreading and includes harmful health advises, wild conspiracy theories, as well as misinformation related to racism. 

Recent machine learning advances have shown significant promise for the detection of misinformation. Examples include approaches that define high level features and use those to describe the distribution of misinformation in a high dimensional space and then train a separating hyperplane in a supervised manner \cite{castillo2011information,kwon2013prominent,yang2012automatic}, as well as  approaches that are based on deep neural networks \cite{ruchansky2017csi,nguyen2017early}. The first set of solutions rely on the extraction of handcrafted features that cannot be optimal for all scenarios. The automatic extraction of features that best explain the underlying misinformation in data presents a significant challenge.  On the other hand, solutions that use low-level features such as those based on deep learning have demonstrated
superior performance in a wide variety of machine learning tasks, however the interpretation of the prediction results, i.e., identification of the key elements that characterize misinformation, is not direct and easily accessible. Even with the use of additional steps such as heat maps \cite{montavon2018methods}, they do not lend themselves to easy interpretation. Both of these challenges become more pronounced in high impact events as interpretability of the prediction result is key for an analyst to explain how true and false information correlate with each other and how they propagate on the complex network of social media \cite{vosoughi2018spread}.

Data driven methods based on latent variable analysis provide the ideal starting point for developing an effective solution that can reliably respond to these complex set of requirements. They have demonstrated great potential for text analysis where interpretable linguistic features are extracted through {\it latent variables} \cite{honkela2010wordica, kolenda2000independent,guacho2018semi}. In this work, we propose a simple yet effective data-driven solution that is based on independent component analysis (ICA) to generate low dimensional representations of tweets with respect to their spatial context. These features are then used to train a classifier to distinguish between reliable and unreliable tweets. It is important to note that the proposed approach is {\em completely data driven} and is easily {\it interpretable} due to the simplicity of the generative model, and {\it does not require a large amount of training samples} in order to achieve a desirable classification error. Another important contribution is the generation of a novel labeled COVID-19 Twitter dataset that has been constructed by subject matter experts and where the labeling generation is based on socio-linguistic criteria. This is an important aspect we would like to  emphasize for datasets used for misinformation detection, and needless to say, in the scenario for COVID-19 as well. 

The remainder of this paper is organized as follows. In Section \ref{MaterialMethods}, we show how we cast our problem using the ICA method and present the classification procedure. Moreover, we show how we collected and constructed the labeled COVID-19 dataset. Section \ref{Experiments} provides the results of the classification procedure, knowledge discovery, and associated discussions. The conclusions and future research directions are presented in Section \ref{Discussion}.

\section{Materials and Methods}\label{MaterialMethods}

\subsection{Independent Component Analysis}
We formulate the problem of detection of unreliable tweets as a blind source separation problem in the following way. Let ${\bf X}\in \mathbb{R}^{d\times V}$ be the observation matrix where $d$ denotes the number of initial feature vectors and $V$ denotes the total number of tweets. For the purpose of this work, we select $d$ to denote the count of the words that appear in the corpus. The noiseless blind source separation model is given by 

\begin{equation}\label{OriginalBSS}
	{\bf X} = {\bf A}{\bf S},
\end{equation}
where ${\bf A}\in \mathbb{R}^{d\times N}$ is the mixing matrix, and ${\bf S}\in \mathbb{R}^{N\times V}$ correspond to the source estimates, 
which in our setting, correspond to the feature vectors. One of the most widely used methods for solving (\ref{OriginalBSS}) is ICA and its basic assumption is that the source signals are statistically independent \cite{AdaliDiversity, hyvarinen2000independent}.

Using the random vector notation, the ICA model is given by ${\bf x}(v) = {\bf A}{\bf s}(v),~~~ v=1,\dots,V, \nonumber$ where $v$ is the sample index, ${\bf s}(v)\in \mathbb{R}^{N}$ are the unknown sources that need to be estimated, and ${\bf x}(v)\in \mathbb{R}^{d}$ are the mixtures. In most practical applications, we have an over-determined system where $d>N$. This can be reduced to the case where $d = N$ using a dimensionality reduction 
technique, typically with principal component analysis (PCA) and an order estimate to determine the signal subspace \cite{wax1985detection,fuicassp2012}.
Under the assumption that the sources $s_n(v),\ 1\leq n\leq N$ in ${\bf s}(v) = [ s_1(v),\dots,s_N(v)]^{\top}$ are statistically independent, the goal in ICA is to estimate a demixing matrix ${\bf W}\in \mathbb{R}^{N\times N}$ to yield maximally independent source estimates ${\bf y}(v) = {\bf W}{\bf x}(v)$. For the rest of the paper, we assume that the samples of each unknown source are independent and identically distributed ,and therefore, to simplify the notation we drop the sample index $v$. A natural cost function to achieve such a separation is mutual information (MI), which is defined as the Kullback-Leibler (KL)-distance between the joint source density and the product of the marginal estimated source densities. Therefore, the MI cost function is given by
\begin{equation}\label{ICAmutualInfo}
	J_{ICA}({\bf W}) = \sum_{n=1}^N H(y_n) - \log|\det({\bf W})| - C,
\end{equation}
where $y_n = {\bf w}_n^{\top} {\bf x}$ and the terms $H(y_n)$ and $H({\bf x})$ are the (differential) entropy of the source estimates and the mixtures, respectively. Though there are many ICA algorithms, in this work, we used the entropy bound minimization (ICA-EBM) algorithm \cite{ICA}, due to the fact that it has shown superior performance in a wide range of applications. 

\subsection{Feature Extraction and Classification Procedure}\label{FeatureExtraction}
The classification process consists of three stages. In the first stage, $N$ tweets are randomly sub-sampled in order to generate training and testing datasets, ${\bf X}_{\rm Train}$ and ${\bf X}_{\rm Test}$ respectively. 

In the second stage, the mean from the initial $d$ feature vectors is removed so all features are centered and PCA is applied to ${\bf X}_{\rm Train}$ using order $N$ such that the data is projected onto an $N-$dimensional sub-space. Then we generate $\hat{\bf X}_{\rm Train}\in \mathbb{R}^{N\times V_{\rm Train}}$ and perform ICA on it. This results a demixing matrix, ${\bf W}_{\rm Train}\in \mathbb{R}^{N\times N}$ that will be used to generate ${\bf Y}_{\rm Train} = {\bf W}_{\rm Train}{\hat{\bf X}}_{\rm Train}$. Note here that ${\bf Y}_{\rm Train}\in \mathbb{R}^{N\times V_{\rm Train}}$ contains the extracted low dimensional features for each training tweet and will be used to train the classification model. 

The testing dataset is generated by removing the mean from each testing feature vector resulting in ${\bf X}_{\rm Test}\in \mathbb{R}^{N\times V_{\rm Test}}$. In order to project the testing data onto a low dimensional feature space, remember that before ICA is performed, PCA is applied to ${\bf X}_{\rm Train}$ using an estimated order $N$. This provides the data reduction matrix ${\bf F}_{\rm Train}$ where ${\bf F}_{\rm Train}$ is formed by the eigenvectors with the first $N$ highest eigenvalues of the corresponding ${\bf X}_{\rm Train}$. An estimate of the mixing matrix ${\bf A}_{\rm Train}$ is computed as ${\hat{\bf A}}_{\rm Train} = ({\bf F}_{\rm Train})^{\dagger}({\bf W}_{\rm Train})^{-1}$, where $(\cdot)^{\dagger}$ denotes the pseudo-inverse of a matrix. Then the testing dataset, ${\bf Y}_{\rm Test}$, is formed as ${\bf Y}_{\rm Test} = ({\hat{\bf A}}_{\rm Train}^{\top}{\hat{\bf A}}_{\rm Train})^{-1}{\hat{\bf A}}_{\rm Train}^{\top}{\bf X}_{\rm Test}$. It is worth mentioning here that for the generation of ${\bf Y}_{\rm Train}$ and ${\bf Y}_{\rm Test}$ we do not take into account the class labels of the training and testing samples. 

In the third stage, we train the classification model using $({\bf Y}_{\rm Train})^{\top}$. Here, the specific form of the classification model is unimportant. But to demonstrate a concrete example, we use Support Vector Machines (SVMs), which have shown reliable  performance in a variety of settings, especially with smaller size datasets \cite{cortes1995support}. Once the classification model has been trained we evaluate its performance by using the unseen data $({\bf Y}_{\rm Test})^{\top}$. For all of the experiments associated to SVM, hyper-parameter optimization and model training and testing is done using a nested cross validation scheme. In addition, the order $N$ is estimated through the same cross validation procedure. In the inner loop, parameters are optimized using grid search selection using $80\%$ of the data to train the model and another $10\%$ for validation. The remaining $10\%$ is held out as a test set to estimate performance after hyperparameter optimization. The outer loop is done five times, corresponding to five folds.

\subsection{Development of Labeled Twitter COVID-19 Dataset}
In constructing our labeled Twitter dataset we initially randomly collected a sample of 282, 201 Twitter users from Canada by using the Conditional Independence Coupling (CIC) method \cite{white2012sampling}. CIC matches the prior distribution of the population, in this case the Canadian general population, ensuring that the sample is balanced for gender, race and age. All tweets posted by these 282,201 people from January 1, 2020 to March 13, 2020 were collected and further analyzed. 

By starting with this large poll of 282,201 tweets we randomly down-sampled 1,600 tweets in order to create a balanced dataset with reliable and unreliable tweets. To eliminate data bias two subject matter experts from our group independently reviewed each tweet to determine whether or not each tweet should be labeled as reliable or unreliable. Tweets are labeled as unreliable if they included content that promotes {\it political bias}, {\it conspiracy}, {\it propaganda}, {\it anger}, or {\it racism} and thus such tweets could affect decision making and create social and political unrest during COVID-19. Due to the complexity of this task, tweets that were not labeled as unreliable by both experts were kept for a second review and finally marked as unreliable if both experts agreed on their decision. This resulted in 280 unreliable tweets. Since the number of reliable tweets is greater than the number of unreliable tweets and thus the final dataset would be highly imbalanced, we randomly down-sampled the reliable class to obtain a balanced dataset. It is worth mentioning that after we down-sampled the reliable class, we also manually checked this class for consistency and validity with respect to reliability. The final dataset consists of 280 reliable and 280 unreliable tweets. 

After the dataset of 560 tweets was compiled, each unreliable tweet was also reviewed for the presence of linguistic attributes that might indicate unreliability, following a two-part process. This process would provide a set of linguistic rules that might be of independent scientific interest but more importantly they can be used to assess the interpretation ability of our classification model. First, each tweet was reviewed for the presence of linguistic characteristics that have been identified in the literature as being indicative of or associated with misinformation, bias, and/or less reliable sources in news media. Second, each tweet was reviewed for the presence of any additional distinguishing linguistic characteristics that appeared to be indicative of unreliability. From the two-part deductive and inductive labeling process, a list of 17 linguistic characteristics was developed. Each tweet was then labeled by hand, indicating any of the 17 characteristics that may have been present. 

Table \ref{table0}, presents the list of the 17 linguistic characteristics, including citations to relevant literature where available, and provides examples of characteristics collected from actual tweets in the dataset.

\begin{table}
	\caption{17 linguistic characteristics identified on the 560 Twitter dataset}
	\label{table0}

	{\footnotesize \begin{tabular}{  m{17em} | m{10cm} } 
			\hline
			{\bf Linguistic Feature}& {\bf Example from Dataset}\\ 
			\hline\hline
			Hyperbolic, intensified, superlative, or emphatic language \cite{baly2018predicting,rashkin2017truth}& e.g., `blame', `accuse', `refuse', `catastrophe', `chaos', `evil' \\ 
			\hline
			Greater use of punctuation and/or special characters \cite{baly2018predicting,perez2017automatic}& e.g., `YA THINK!!?!!?!', `Can we PLEASE stop spreading the lie that Coronavirus is super super super contagious? It's not. It has a contagious rating of TWO' \\ 
			\hline
			Strongly emotional or subjective language \cite{baly2018predicting,vosoughi2018spread,wilson2005recognizing,rashkin2017truth,pennebaker2001linguistic} & e.g., `fight', `danger', `hysteria', `panic', `paranoia', `laugh', `stupidity' or other words indicating fear, surprise, alarm, anger, and so forth \\ 
			\hline
			Greater use of verbs of perception and/or opinion \cite{perez2017automatic}& e.g., `hear', `see', `feel', `suppose', `perceive', `look', `appear', `suggest', `believe', `pretend'\\ 
			\hline
			Language related to death and/or war \cite{torabi} & e.g., `martial law', `kill', `die', `weapon', `weaponizing'\\ 
			\hline
			Greater use of proper nouns \cite{horne2017just} & e.g., `USSR lied about Chernobyl. Japan lied about Fukushima. China has lied about Coronavirus. Countries lie. Ego, global'\\ 
			\hline
			Shorter and/or simpler, language \cite{horne2017just} & e.g., `\#Iran just killed 57 of our citizens. The \#coronavirus is spreading for Canadians   Our economy needs support.'\\ 
			\hline
			Hate speech \cite{torabi} and/or use of racist or stereotypical language& e.g., `foreigners', `Wuhan virus', reference to Chinese people eating cats and dogs\\ 
			\hline
			First and second person pronouns \cite{rashkin2017truth,perez2017automatic} & e.g., `I', `me', `my', `mine', `you', `your', `we', `our'\\ 
			\hline
			Direct falsity claim and/or a truth claim \cite{baly2018predicting}& e.g., `propaganda', `fake news', `conspiracy', `claim', `misleading', `hoax'\\ 
			\hline
			Direct health claim & e.g., `cure', `breakthrough', posting infection statistics\\ 
			\hline
			Repetitive words or phrases \cite{horne2017just}& e.g., `Communist China is lying about true extent of Coronavirus outbreak - If Communist China doesn’t come clean'\\ 
			\hline
			Mild or strong expletives, curses, slurs, or other offensive terms & e.g., `bitch', `WTF', `dogbreath', `Zombie homeless junkies', `hell', `screwed'\\ 
			\hline
			Language related to religion & e.g., `secular', `Bible'\\ 
			\hline
			Politically biased terms & e.g., `MAGA', `MAGAt', `genetic engineer Hillary', `Chinese regime', `deep state', `Free Market Fundamentalists', `Communist China', `Nazi'\\ 
			\hline
			Language related to financial or economic impact, money/costs, or the stock market & e.g., `THE STOCK MARKET ISN'T REAL THE ECONOMY ISN'T REAL THE CORONAVIRUS ISN'T REAL FAKE NEWS REEEEEEEEEEEEEEEEE'\\ 
			\hline
			Language related to the Trump presidential election, campaign, impeachment, base, and rallies & e.g., `What you are watching with the CoronaVirus has been planned and orchestrated. We are 8 months from the next Presidential elections'\\
	\end{tabular}}
\end{table}

\section{Experimental Results} \label{Experiments}
The study examines (Sections \ref{Classificationresults} and \ref{Sparsity}) classification performance as well as interpretability properties of ICA (Section \ref{KnowledgeDiscovery})  using our new labeled Twitter dataset. The relatively small size of the dataset is in line with
one of the aspects of our work. To construct the data matrix ${\bf X}$ we pre-process the content of each tweet from out labeled dataset using tokenization. Besides, stop-words and punctuation were removed as well as all words are converted to lower case characters. After pre-processing, each entry in ${\bf X}\in \mathbb{R}^{2569\times 560}$, initially represents the frequency of a word from the corpus in tweet $v$ and we further process the raw frequencies with tf-idf weighting. The order $N$ has been selected through cross-validation as described in (\ref{FeatureExtraction}) and the optimal value is $N=50$. 

\subsection{Classification Analysis}\label{Classificationresults}

We demonstrate the performance of our ICA based classification model in terms of accuracy (percentage of correctly classified tweets), sensitivity (percentage of all unreliable tweets that are correctly
predicted as unreliable), precision (percentage of tweets predicted as unreliable out of
all tweets predicted as unreliable), and  F-1 score (harmonic mean of precision and sensitivity). Due to the superior performance in many natural language processing techniques we compare our method with three commonly used "deep" classification algorithms: Long Short Term Memory (LSTM) \cite{hochreiter1997long}; its derivative, Bidirectional Long Short Term Memory (BiLSTM) \cite{huang2015bidirectional}; and Bidirectional Encoder Representations from Transformers (BERT) \cite{BERT}. All three methods are based on neural models and have proved to be effective in sequence classification tasks. LSTM models, which are an improvement over Recurrent Neural Networks (RNNs), are more effective than the latter for classification of text which multiple sentences. This is due to the presence of a learn-able `forget' parameter which helps the model to forget/discard non-critical information. BiLSTMs, as opposed to regular LSTMs, unroll in two directions, which helps them to retain crucial information of the `past' (preceding a particular word) as well as the `future' (succeeding a particular word). BERT classification models are pre-trained on unsupervised language modeling tasks before being fine-tuned for a downstream classification task. We ran an experiment using basic LSTM, basic BiLSTM, and BERT$_{\small \textsc{BASE}}$, and tracked the performance of those models in our dataset. We made a 70:30 (train: validation) split on the dataset. Both the LSTM and BiLSTM models had an embedding layer of 64 dimensions. The `maxlength' parameter of all three models was chosen to be 30 (meaning it will discard any word post the 30th word of the sequence). Our experiments showed that this parameter did not have much effect for our dataset, since the data is twitter feeds which have a character limit in itself. For LSTM and BiLSTM, the `dropout' parameter had a value of 0.5. We used 64 LSTM cells, and 64 BiLSTM cells for our models. Both the LSTM and BiLSTM models had a near-perfect score, when evaluated on the training set (0.995 F-1 for both LSTM and BiLSTM). However, the performance dipped drastically when evaluated on the unseen test set. This proves that these two models have the problem of `over-fitting'. BERT also scored well during training (0.990 F-1), but demonstrated much lower generalization error. In addition we show how our proposed classification method is affected by the choice of different kernels in SVM. From Table \ref{table1}, we see that the proposed ICA based classification model with Gaussian kernel performs the best in terms of the accuracy, sensitivity, and F1 score. We note that our method is able to achieve this performance under the scenario of a small dataset due to its model simplicity and thus demonstrating the fact that latent variable based approaches do not require a large amount of training samples in order to achieve a desirable performance. On the other hand, LSTM and BiLSTM, had to train 188,833 and 198,017 parameters respectively and due to the small training dataset they model the training data too well and thus the perform poorly with unseen data. While BERT generalized well to unseen data, the model requires fine-tuning of 110M parameters, and was the most computationally expensive to train (2 min 6 seconds on a NVIDIA Tesla K80 GPU). On the other hand, the proposed method requires 18.58 seconds where 2.96 seconds correspond to a single ICA run and 15.61 seconds correspond to the training time for the SVM classifier. For the proposed method we performed our experiments on a 4 Core Intel i7 1.9 GHz. 

\begin{table}
	\caption{Classification performance for different methods.}
	\label{table1}
	
	\centering
	{\footnotesize\begin{tabular}{|c || c | c | c | c |}
			\hline
			Method & Accuracy & Sensitivity & Precision & F-1\\
			\hline\hline
			
			{\bf DNN} &  &  &  & \\
			
			LSTM & 0.515 & 0.516 & 0.516 & 0.515\\  
			
			BiLSTM & 0.574 & 0.579 & 0.575 & 0.574 \\ 
			
			BERT$_{\small \textsc{BASE}}$ & {\bf 0.875} & {\bf 0.900} & {\bf 0.847} & {\bf 0.873} \\ 
			
			\hline\hline
			{\bf ICA} &  &  &  & \\
			
			SVM/Gaussian & {\bf 0.812} & {\bf 0.763} & {\bf 0.859} & {\bf 0.803}\\
			
			SVM/RBF & 0.796 & 0.768 & 0.824 & 0.791 \\ 
			
			SVM/Polynomial & 0.794 & 0.762 & 0.8267 & 0.784\\
			
			\hline\hline
			{\bf Sparse ICA} &  &  &  & \\
			
			$\lambda = 100$ & {\bf 0.691} & {\bf 0.679} & 0.744 & 0.644 \\ 
			
			$\lambda = 1000$ & 0.687 & 0.613 & {\bf 0.834} & {\bf 0.672}\\
			\hline
	\end{tabular}}
\end{table}

\subsection{Sparsity-aware ICA Classification}\label{Sparsity}

Due to the nature of the data (word counts/tf-idf), sparsity, is a type of property of the data that could be taken into account during the feature generation as described in (\ref{FeatureExtraction}). However, the main question that arises in our application is whether or not taking this type of diversity into account will yield better classification performance. In many practical applications, it has been shown that if sparsity is incorporated into the ICA model it can relax the independence assumption, resulting in an improvement in the overall separation performance \cite{boukouvalas2018sparsity,boukouvalas2017enhancing}. Although this may be true in terms of the separation power of the model, it is not true in terms of the classification accuracy. We investigate this by replacing ICA-EBM with SparseICA-EBM \cite{boukouvalas2018sparsity,boukouvalas2017enhancing} during the feature extraction and classification procedure as described in (\ref{FeatureExtraction}) and by using a Gaussian kernel for SVM. SparseICA-EBM is based on a unified mathematical framework that enables direct control over the influence of statistical independence and sparsity. This is implemented by penalizing (\ref{ICAmutualInfo}) by an $\ell^1$ regularization term, thus, enabling a direct exploitation of sparsity for each source individually through a single sparsity parameter $\lambda$. We should note that if $\lambda = 0$ then SparseICA-EBM is equivalent to ICA-EBM and as $\lambda$ increases model favors more sparsity than statistical independence during the decomposition. 

Table \ref{table1} shows the performance of the proposed model when ICA-EBM is replaced by SparseICA-EBM using a Gaussian kernel and for different values of the parameter ${\lambda}$. We observe that as $\lambda$ increases the classification accuracy significantly decreases revealing that imposing sparsity to the underlying features does not benefit the classification accuracy. The main reason for this behavior is due to the fact that SparseICA-EBM produces highly correlated features and as expected it negatively affects the discriminating power of the classification model.

\subsection{Knowledge Discovery}\label{KnowledgeDiscovery}

An important property of ICA is that it can also provide linguistic interpretations through the estimated mixing matrix ${\hat{\bf A}}$. Each row of ${\hat{\bf A}}$ represents the weights for the estimated sources of the initial features. The values of the weights can reveal relationships between certain characteristics of a given set of tweets. To demonstrate this further we sort the magnitudes of the weights on each estimated column of mixing matrix ${\hat{\bf A}}$ and the corresponding features used in the analysis. Note that in our application the initial features are the word counts from the corpus that we created using the proposed Twitter dataset. 

From Table \ref{ConceptsTrue}, we see the most representative words that belong to each extracted feature that is associated with the reliable tweets. From Table \ref{ConceptsTrue}, it can be seen that each latent concept contains words with different conceptual meaning. For instance, Feature 1, contains words that are related to the cancellation of the Olympic games that are happening in Tokyo due to the COVID-19. Feature 5, is populated by words with similar contextual characteristics to ``New York" and are related to the number of cases in this state as well as the declaration of New York as an emergency state. 

On the other hand, Table (\ref{ConceptsFake}), contains extracted features that are related to unreliable tweets. For instance, Feature 2, contains the word Pirbright Institute as well as all the words that are associated with the ``rumor" that coronavirus is designed in Pirbright Institute, that Bill Gates patent COVID-19, and that there is vaccine. Feature 5, contains politically biased and offensive terms, while feature 4, contains subjective/emotional language that is related to the stock market. Feature 3, is a direct reference to falsity and contains words that link to dangerous health advises regarding the virus. Finally, feature 1 links hyperbolic, intensified, or emphatic language with direct reference to falsity terms and political terms as presented in Table \ref{table0}. This reveals that extracted features that are associated with unreliable tweets are in agreement with most of the 17 linguistic characteristics that are presented in Table \ref{table0}.

To conclude, we note that each feature obtained through the exploitation of statistical independence carries semantic linguistic information providing a basis for knowledge discovery during COVID-19.  

\begin{table}
	\caption{The 15 most representative words for five extracted features that are related to reliable tweets.}
	\centering{
	{\footnotesize\begin{tabular}{c | c | c | c | c }
			\hline
			{\bf Feature 1} & {\bf Feature 2} & {\bf Feature 3} & {\bf Feature 4} & {\bf Feature 5} \\
			\hline\hline
			\#coronavirus 	& positive 		& time 		& conference 	& new \\  
			
			cases 			& covid			& \#covid  	& community 	& state \\ 
			
			\#covid 		& tested		& social 	& weeks 		& york \\
			
			th  			& tests 		& distancing & few 			& cases\\
			
			iran 			& \#coronavirus	& systems	& best  		& emergency \\
			
			confirmed 		& jazz 			& emergency & bioinformatics  & declares \\
			
			virus			& player 		& get		& \#bcc 		& death \\
			
			canada 			& patient 		& new 		& postpone 		& cuomo \\
			
			ontario			& season 		& down		& registration 	& washington \\
			
			olympic 		& nba 			& \#coronavirus	& post  	& number\\
			
			\#tokyo 		& utah 			& wuhan		& decided  		& times\\
			
			man 			& test 			& shut		& full  		& reported\\
			
			road 			& transit 			& trump	& figure  		& reports\\
			
			cancellation 	& toronto 			& give	& news  		& first\\
			
			\#ioc 			& amp 			& amp		& covid  		& deaths\\
			\hline\hline		
	\end{tabular}}

	\label{ConceptsTrue}}
\end{table}

\begin{table}
		\caption{The 15 most representative words for five extracted features that are related to unreliable tweets.}
	\centering{
	{\footnotesize \begin{tabular}{c | c | c | c | c }
			\hline
			{\bf Feature 1} & {\bf Feature 2} & {\bf Feature 3} & {\bf Feature 4} & {\bf Feature 5} \\
			\hline\hline
			trump 		& pirbright 	& inside 		& choose 	& bannon \\  
			
			spreading	& institute		& water  		& fear 		& steve \\ 
			
			response 	& owns			& ayatollahs 	& real 		& turdball \\
			
			people 		& gates 		& solutions 	& market 	& sowing \\
			
			\#coronavirus & funded		& viruses		& stock 	& dissension \\
			
			face		& patent 		& dip 			& abuse 	& ranks \\
			
			trying		& bill 			& anus			& deserves & casting \\
			
			conspiracy 	& cdc 			& bottle 		& life 		& fictitious \\
			
			rank		& blocking 		& cleric		& appearing & net \\
			
			pernicious	& limits 		& spit			& coaching & michelle \\
			
			dodge	 	& drugmakers 	& blessings		& elections & churchill \\
			
			accountability	& charge 		& joke			& rise 		& moment \\
			
			theories 	& gop 			& oil			& everything & democratic \\
			
			flooding 	& outbreak 		& sleep			& false 	& \#kag \\
			
			chaos		& covid 		& drinking		& help 		& \#maga \\
			\hline\hline		
	\end{tabular}}

	\label{ConceptsFake}}
\end{table}

\section{Discussion}\label{Discussion}

In this work, we present a data-driven solution that is based on ICA such that detection of misinformation and knowledge discovery can be achieved jointly. In addition, we present a novel labeled COVID-19 Twitter dataset that will enable the research community to study the phenomenon of misinformation during the COVID-19 high impact event.

The success of the proposed method raises several interesting questions both in terms of the use of a latent variable method for the task as well as a computational socio-linguistic perspective that can be explored in future work. In terms of latent variable methods, it would be of high interest to investigate which types of properties, such as, non-linearity, non-negativeness, sample-to-sample correlation, could affect the classification performance. Moreover, for this work, we considered a single featurization method which is based on the bag of word model in order to generate ${\bf X}$. Since choosing which featurization method is the most appropriate for a particular machine learning task is highly non-trivial, an interesting direction is to simultaneously use sophisticated featurization methods. This can be achieved by considering a more advanced model such as Independent Vector Analysis \cite{kim2006independent, AdaliDiversity, boukouvalas2015efficient}, that will enable us to quantify the importance of each featurization method during a high impact event with respect to the quality of the decomposition as well as the affect on the classification performance. Finally, in terms of a computational socio-linguistic perspective, the development of validation techniques for the extracted features and how they can be used to answer questions that record the behavior and interactions of individuals in virtual worlds is a significant research direction and deserves further investigation.

\subsection*{Data Availability}
The labeled COVID-19 Twitter dataset can be found at\ \url{https://zoisboukouvalas.github.io/Code.html}
\subsection*{Acknowledgments}
We thank Dr. Kenton White, Chief Scientist at Advanced Symbolics Inc, for providing the initial Twitter dataset.

\bibliographystyle{unsrt}
\bibliography{references}

\begin{thebibliography}{10}

\bibitem{castillo2011information}
Carlos Castillo, Marcelo Mendoza, and Barbara Poblete.
\newblock Information credibility on twitter.
\newblock In {\em Proceedings of the 20th international conference on World
  wide web}, pages 675--684. ACM, 2011.

\bibitem{kwon2013prominent}
Sejeong Kwon, Meeyoung Cha, Kyomin Jung, Wei Chen, and Yajun Wang.
\newblock Prominent features of rumor propagation in online social media.
\newblock In {\em 2013 IEEE 13th International Conference on Data Mining},
  pages 1103--1108. IEEE, 2013.

\bibitem{yang2012automatic}
Fan Yang, Yang Liu, Xiaohui Yu, and Min Yang.
\newblock Automatic detection of rumor on sina weibo.
\newblock In {\em Proceedings of the ACM SIGKDD Workshop on Mining Data
  Semantics}, page~13. ACM, 2012.

\bibitem{ruchansky2017csi}
Natali Ruchansky, Sungyong Seo, and Yan Liu.
\newblock Csi: A hybrid deep model for fake news detection.
\newblock In {\em Proceedings of the 2017 ACM on Conference on Information and
  Knowledge Management}, pages 797--806. ACM, 2017.

\bibitem{nguyen2017early}
Tu~Ngoc Nguyen, Cheng Li, and Claudia Nieder{\'e}e.
\newblock On early-stage debunking rumors on twitter: Leveraging the wisdom of
  weak learners.
\newblock In {\em International Conference on Social Informatics}, pages
  141--158. Springer, 2017.

\bibitem{montavon2018methods}
Gr{\'e}goire Montavon, Wojciech Samek, and Klaus-Robert M{\"u}ller.
\newblock Methods for interpreting and understanding deep neural networks.
\newblock {\em Digital Signal Processing}, 73:1--15, 2018.

\bibitem{vosoughi2018spread}
Soroush Vosoughi, Deb Roy, and Sinan Aral.
\newblock The spread of true and false news online.
\newblock {\em Science}, 359(6380):1146--1151, 2018.

\bibitem{honkela2010wordica}
Timo Honkela, Aapo Hyv{\"a}rinen, and Jaakko~J V{\"a}yrynen.
\newblock Word{ICA}—emergence of linguistic representations for words by
  independent component analysis.
\newblock {\em Natural Language Engineering}, 16(3):277--308, 2010.

\bibitem{kolenda2000independent}
Thomas Kolenda, Lars~Kai Hansen, and Sigurdur Sigurdsson.
\newblock Independent components in text.
\newblock In {\em Advances in Independent Component Analysis}, pages 235--256.
  Springer, 2000.

\bibitem{guacho2018semi}
Gisel~Bastidas Guacho, Sara Abdali, Neil Shah, and Evangelos~E Papalexakis.
\newblock Semi-supervised content-based detection of misinformation via tensor
  embeddings.
\newblock In {\em 2018 IEEE/ACM International Conference on Advances in Social
  Networks Analysis and Mining (ASONAM)}, pages 322--325. IEEE, 2018.

\bibitem{AdaliDiversity}
T.~Adal{\i}, M.~Anderson, and Geng-Shen Fu.
\newblock Diversity in independent component and vector analyses:
  Identifiability, algorithms, and applications in medical imaging.
\newblock {\em Signal Processing Magazine, IEEE}, 31(3):18--33, May 2014.

\bibitem{hyvarinen2000independent}
Aapo Hyv{\"a}rinen and Erkki Oja.
\newblock Independent component analysis: algorithms and applications.
\newblock {\em Neural networks}, 13(4):411--430, 2000.

\bibitem{wax1985detection}
Mati Wax and Thomas Kailath.
\newblock Detection of signals by information theoretic criteria.
\newblock {\em IEEE Transactions on Acoustics, Speech, and Signal Processing},
  33(2):387--392, 1985.

\bibitem{fuicassp2012}
G.-S. Fu, H.~Li, M.~Anderson, X.-L. Li, and T.~Adal{\i}.
\newblock Order selection for dependent samples using entropy rate.
\newblock In {\em Proc. IEEE Int.~Conf.~Acoust., Speech, Signal Processing
  (ICASSP)}, pages 2161--2164, Kyoto, Japan, March 2012.

\bibitem{ICA}
X.~{Li} and T.~{Adali}.
\newblock Independent component analysis by entropy bound minimization.
\newblock {\em IEEE Transactions on Signal Processing}, 58(10):5151--5164, Oct
  2010.

\bibitem{cortes1995support}
Corinna Cortes and Vladimir Vapnik.
\newblock Support-vector networks.
\newblock {\em Machine learning}, 20(3):273--297, 1995.

\bibitem{white2012sampling}
Kenton White, Guichong Li, and Nathalie Japkowicz.
\newblock Sampling online social networks using coupling from the past.
\newblock In {\em 2012 IEEE 12th International Conference on Data Mining
  Workshops}, pages 266--272. IEEE, 2012.

\bibitem{baly2018predicting}
Ramy Baly, Georgi Karadzhov, Dimitar Alexandrov, James Glass, and Preslav
  Nakov.
\newblock Predicting factuality of reporting and bias of news media sources.
\newblock {\em arXiv preprint arXiv:1810.01765}, 2018.

\bibitem{rashkin2017truth}
Hannah Rashkin, Eunsol Choi, Jin~Yea Jang, Svitlana Volkova, and Yejin Choi.
\newblock Truth of varying shades: Analyzing language in fake news and
  political fact-checking.
\newblock In {\em Proceedings of the 2017 Conference on Empirical Methods in
  Natural Language Processing}, pages 2931--2937, 2017.

\bibitem{perez2017automatic}
Vernica Perez-Rosas, Bennett Kleinberg, Alexandra Lefevre, and Rada Mihalcea.
\newblock Automatic detection of fake news.
\newblock {\em arXiv preprint arXiv:1708.07104}, 2017.

\bibitem{wilson2005recognizing}
Theresa Wilson, Janyce Wiebe, and Paul Hoffmann.
\newblock Recognizing contextual polarity in phrase-level sentiment analysis.
\newblock In {\em Proceedings of human language technology conference and
  conference on empirical methods in natural language processing}, pages
  347--354, 2005.

\bibitem{pennebaker2001linguistic}
James~W Pennebaker, Martha~E Francis, and Roger~J Booth.
\newblock Linguistic inquiry and word count: Liwc 2001.
\newblock {\em Mahway: Lawrence Erlbaum Associates}, 71(2001):2001, 2001.

\bibitem{torabi}
The language gives it away: How an algorithm can help us detect fake news.
\newblock
  \url{https://theconversation.com/the-language-gives-it-away-how-an-algorithm-can-help-us-detect-fake-news-120199},
  2019.

\bibitem{horne2017just}
Benjamin~D Horne and Sibel Adali.
\newblock This just in: fake news packs a lot in title, uses simpler,
  repetitive content in text body, more similar to satire than real news.
\newblock In {\em Eleventh International AAAI Conference on Web and Social
  Media}, 2017.

\bibitem{hochreiter1997long}
Sepp Hochreiter and J{\"u}rgen Schmidhuber.
\newblock Long short-term memory.
\newblock {\em Neural computation}, 9(8):1735--1780, 1997.

\bibitem{huang2015bidirectional}
Zhiheng Huang, Wei Xu, and Kai Yu.
\newblock Bidirectional lstm-crf models for sequence tagging.
\newblock {\em arXiv preprint arXiv:1508.01991}, 2015.

\bibitem{BERT}
Jacob Devlin, Ming{-}Wei Chang, Kenton Lee, and Kristina Toutanova.
\newblock {BERT:} pre-training of deep bidirectional transformers for language
  understanding.
\newblock {\em CoRR}, abs/1810.04805, 2018.

\bibitem{boukouvalas2018sparsity}
Zois Boukouvalas, Yuri Levin-Schwartz, Vince~D Calhoun, and T{\"u}lay Adal{\i}.
\newblock Sparsity and independence: Balancing two objectives in optimization
  for source separation with application to fmri analysis.
\newblock {\em Journal of the Franklin Institute}, 355(4):1873--1887, 2018.

\bibitem{boukouvalas2017enhancing}
Zois Boukouvalas, Yuri Levin-Schwartz, and T{\"u}lay Adal{\i}.
\newblock Enhancing ica performance by exploiting sparsity: Application to fmri
  analysis.
\newblock In {\em 2017 IEEE International Conference on Acoustics, Speech and
  Signal Processing (ICASSP)}, pages 2532--2536. IEEE, 2017.

\bibitem{kim2006independent}
Taesu Kim, Torbj{\o}rn Eltoft, and Te-Won Lee.
\newblock Independent vector analysis: An extension of ica to multivariate
  components.
\newblock In {\em International conference on independent component analysis
  and signal separation}, pages 165--172. Springer, 2006.

\bibitem{boukouvalas2015efficient}
Zois Boukouvalas, Geng-Shen Fu, and T{\"u}lay Adal{\i}.
\newblock An efficient multivariate generalized gaussian distribution
  estimator: Application to iva.
\newblock In {\em 2015 49th Annual Conference on Information Sciences and
  Systems (CISS)}, pages 1--4. IEEE, 2015.

\end{thebibliography}

\end{document}